# Outlier Detection using Improved Genetic K-means

M. H. Marghny
Computer ScienceDepartment,
Faculty of Computer and
Information, Assiut University,
Egypt.

Ahmed I. Taloba
Computer ScienceDepartment,
Faculty of Computer and
Information, Assiut University,
Egypt.

## ABSTRACT
The outlier detection problem in some cases is similar to the classification problem. For example, the main concern of clustering-based outlier detection algorithms is to find clusters and outliers, which are often regarded as noise that should be removed in order to make more reliable clustering.

In this article, we present an algorithm that provides outlier detection and data clustering simultaneously. The algorithmimprovesthe estimation of centroids of the generative distribution during the process of clustering and outlier discovery. The proposed algorithm consists of two stages. The first stage consists of improved genetic k-means algorithm (IGK) process, while the second stage iteratively removes the vectors which are far from their cluster centroids.

## General Terms
Data Mining.

## 1. INTRODUCTION
Data mining, in general, deals with the discovery of non-trivial, hidden and interesting knowledge from different types of data. With the development of information technologies, the number of databases, as well as their dimension and complexity, grow rapidly. It is necessary what we need automated analysis of great amount of information. The analysis results are then used for making a decision by a human or program. One of the basic problems of data mining is the outlier detection.

An outlier is an observation of the data that deviates from other observations so much that it arouses suspicions that it was generated by a different mechanism from the most part of data [1]. Inlier, on the other hand, is defined as an observation that is explained by underlying probability density function. This function represents probability distribution of main part of data observations [2].

Many data-mining algorithms manipulate outliers as a side-product of clustering algorithms. However these techniques define outliers as points, which do not lie in clusters. Thus, the techniques implicitly define outliers as the background noise in which the clusters are embedded. Another class of techniques defines outliers as points, which are neither a part of a cluster nor a part of the background noise; rather they are specifically points which behave very differently from the norm [3]. Some noisy points may be far away from the data points, whereas the others may be close. The faraway noisy points would affect the result more significantly because they are more different from the data points. It is desirable to identify and remove the outliers, which are far away from all the other points in cluster [4]. So, to improve the clustering such algorithms use the same process and functionality to solve both clustering and outlier discovery [2].

In this paper, we propose a clustering-based technique to identify outliers and simultaneously produce data clustering. Proposed outlier detection process at the same time is effective for extracting clusters and very efficient in finding outliers.

## 2. IMPROVED GENETIC K-MEANS (IGK)
IGK is an efficient clustering algorithm to handle large scale data, which can select initial clustering center purposefully using Genetic algorithms (GAs), reduce the sensitivity to isolated point, avoid dissevering big cluster, and overcome deflexion of data in some degree that caused by the disproportion in data partitioning owing to adoption of multi-sampling [5].

The mean steps of proposed algorithm can be summarized as follows:

**Algorithm**: Improved Genetic K-means (S, k), S = {x₁, x₂,…,xₙ}.

**Input**: The number of clusters K'(K'> K) and a dataset containing n objects xᵢ.

**Output**: A set of k clusters Cⱼ that minimize the squared-error criterion.

Begin
1. Multiple sub-samples {S₁, S₂, ...,Sⱼ};
2. For m = 1 to j do
   Genetic K-means(Sₘ, K'); //executing Genetic K-means, produce K' clusters and j groups.
3. Compute$J_c(m) = \sum_{j=1}^{kr} \sum_{X_i \in c_j} |X_i - Z_j|^2$;
4. Choose min{Jᵣ} as the refined initial points Zⱼ , j ∈ [1, K'];
5. Genetic K-means(S, K'); //executing Genetic K-means again with chosen initial, producing K'mediods.
6. Repeat
   Combining two near clusters into one cluster, and recalculate the new center generated by two centers merged.
7. Until the number of clusters reduces into k //Merging (K' +K)
End

## 3. OUTLIER DETECTION METHODS
In outlier detection methods based on clustering, outlier is defined to be an observation that does not fit to the overall clustering pattern [6]. The ability to detect outliers can be improved using a combined perspective of outlier detection and clustering. Some clustering algorithms [7-16] handle outliers as special observations, but their main concern is clustering the dataset, not detecting outliers. The following techniques have been proposed to detect outliers:

**Outlier Detection using Indegree Number (ODIN)**[15] is a local density-based outlier detection algorithm. Local density based scheme can be used in cluster thinning. Outlier removal algorithm can remove vectors from the overlapping regions between clusters, if the assumption holds that the regions are of relatively low density. Higher density is found near the cluster centroid. An obvious approach to use outlier rejection in the cluster thinning is as follows: (i) eliminate outliers (ii) cluster the data using any method.

In ODIN, the outliers are defined using a k-nearest neighbour (kNN) graph, in which every vertex represents a data vector, and the edges are pointers to neighbouring k vectors. The weight of an edge $e_{ij}$ is $||x_i - x_j||$. In ODIN, the outlyingness of $x_i$ is defined as:

$$O_i = \frac{1}{ind(x)+1}$$

Where $ind(x_i)$ is the indegree of the vertex $x_i$, i.e. the number of edges pointing to $x_i$. In the first step of ODIN, a kNN graph is created for the dataset X. Then, each vertex is visited to check if its outlyingness is above threshold T.
An algorithm shows the ODIN method as follows:

**Algorithm ODIN+K-means** (k, T):

```
Begin
{ind(xi)|i = 1, ...,N}←Calculate kNN graph
fori←1,...,N do
          Oi←1/(ind(xi) + 1)
          ifOi> T then
                     X←X \ {xi}
          end if
end for
(C,P)←K-means(X) //C is the centers and P is the partition.
End
```

**Outlier removal clustering (ORC)**[16], it consists of two consecutive stages, which are repeated several times. In the first stage, K-means algorithm is performed until convergence, and in the second stage, the outlyingness factor is assigned for each vector. Factor depends on its distance from the cluster centroid. Then algorithm iterations start, with first finding the vector with maximum distance to the partition centroid $d_{max}$.

$$d_{max} = \max_i \left\| x_i - c_{pi} \right\|, i = 1,...,N$$

The outlyingness of a vector $x_i$ is defined as follows:

$$O_i = \frac{\left\| x_i - c_{pi} \right\|}{d_{max}}$$

An algorithm shows the ORC method as follows:
**Algorithm ORC** (I, T)

```
Begin
C ← Run K-means with multiple initial solutions, pick best C
for j← 1,...,I do //I is no of iterations
          dmax← maxi{||xi – cpi||}
          for i ← 1,...,N do
                     Oi = ||xi–cpi||/dmax
                     ifOi> T then
                                X ← X \ {xi}
                     end if
          end for
          (C,P) ← K-means(X,C) //C is the centers and P is the
          partition.
end for
End
```

## 4. PROPOSED METHOD
The ability to detect outliers can be improved using a combined perspective from outlier detection and cluster identification. Unlike the traditional clustering-based methods, the proposed algorithm provides much efficient outlier detection and data clustering capabilities in the presence of outliers. This approach is based on filtering of the data after clustering process. The purpose of our method is not only to produce data clustering but at the same time to find outliers from the resulting clusters.

The algorithm of our outlier detection method is divided into two stages. The first stage carried out using IGK process and the second stage removing outliers according to a chosen threshold.

So, the proposed method is like the ORC method except in the first stage, ORC was applying K-means. But K-means suffers from some drawbacks such as K-means is sensitive to initial choice of cluster centers, the clustering can be very different by starting from different centers, K-means can't deal with massive data, K-means is sensitive with respect to outliers and noise. Hence, we employed IGK instead K-means which has many advantages as we described before.

In our method, the outliers are defined using outlyingness factor that assigned for each vector. Factor depends on its distance from the cluster centroid. Outlyingness factors for each vector are defined as follows:

$$O_i = \frac{\left\| x_i - c_{pi} \right\|}{d_{max}}$$

Where $x_i$ is the vector and $d_{max}$ is the maximum distance between vector $x_i$ and the partition centroid $c_{pi}$ and defined as follows:

$$d_{max} = \max_i \left\| x_i - c_{pi} \right\|, i = 1,...,N$$

We see that all outlyingness factors of the dataset are normalized to the scale [0, 1]. The greater value, the more likely the vector is an outlier. An example of dataset clustered in three clusters and calculated outlyingness factors is shown in Figure 1.

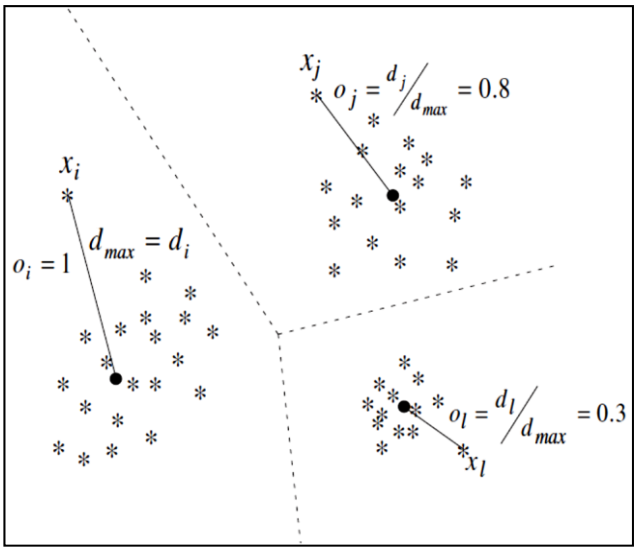

**Fig 1: Example of outlyingness factors.**

The proposed algorithm can be summarized as follows:

**Algorithm Outlier Detection** (I, T)

```
Begin
C ←IGK
for j← 1,...,I do //I is no of iterations
        d_max← max_i{||x_i − c_pi||}
        for i ← 1,...,N do
                O_i = ||x_i−c_pi||/d_max
                ifO_i> T then
                        X ← X \ {x_i}
                end if
        end for
        (C,P) ←IGK(X,C) //C is the centers and P is the
        partition.
end for
End
```

The principle of outliers removing depends on the difference between threshold T and outlyingness factors $O_i$ if $O_i$> T, the outlier is removed from the dataset. Threshold is set by user in range between 0 and 1. At the end of each iteration, IGK is run with the previous C as the initial codebook, so new solution will be a fine-tuned solution for the reduced dataset. By setting the threshold to T < 1, at least one vector is removed. Thus, increasing the number of iterations and decreasing the threshold will in effect remove more vectors from the dataset, possibly all vectors.

The performance of proposed method for 10 iterations with threshold 0.9 is illustrated in Figure 2. In the original data, there are the most distant vectors from centroids, some of such furthest vectors are labeled by arrows in Figure 2 on the left. The algorithm proceeds by removing the furthest vectors from all partitions.Figure 2 on right demonstrates the resulting data after the algorithm.

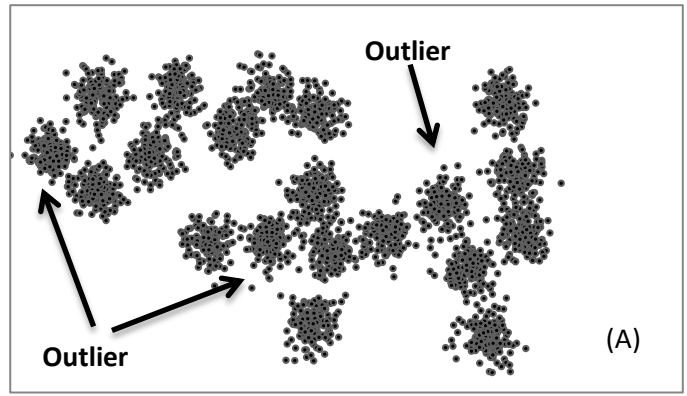

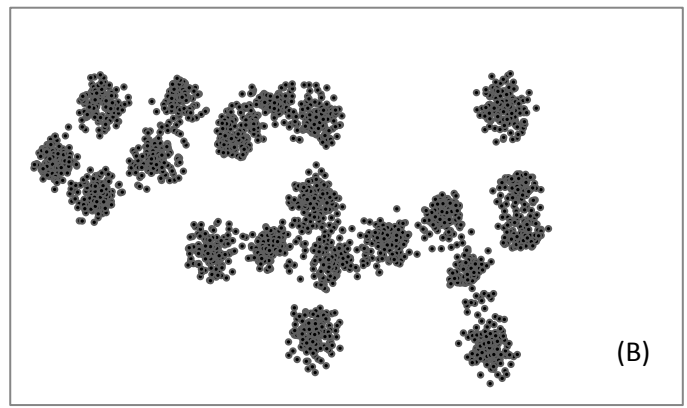

**Fig 2: Example of proposed method. Original dataset (A) and after10 iterations (B) with threshold T = 0.9.**

## 5. EXPERIMENTS

We run experiments on three synthetic datasets denoted as A1, A2 and A3 [17], which are shown in Figure 3 and summarized in Table 1 with threshold equal to 0.9 and number of iterations equal to 10, where N is number of examples and M is no of clusters.

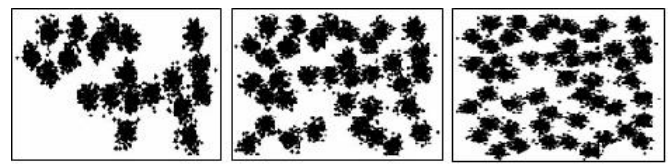

**Fig3: Datasets, A1 (left), A2 (center) and A3 (right).**

The mean square error (MSE) used to differentiate the difference between outlier points. The distance between the estimated centroids and the true centroids has been calculated to computes the error.

**Table 1. Summary of the datasets.**

| Dataset | N | M |
|---------|------|----|
| A1 | 3000 | 20 |
| A2 | 5250 | 35 |
| A3 | 7500 | 50 |

Figure 4, shows that the error is decreasing very fast when the threshold becomes bigger, thus removing more and more number of objects implies to decrease the distance between the points and cluster centroids.

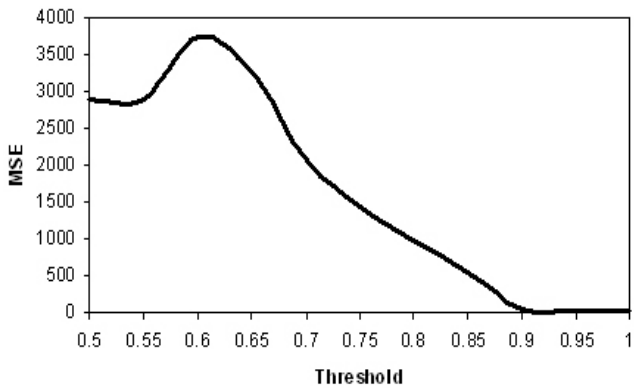

**Fig 4: Comparison between the threshold (T) and MSE**

In Table 2, we show the smallest MSE between original centroids and those obtained by using K-means, ORC and proposed method. It is obvious that the MSE obtained by the proposed method for all data set (A1 14.77, A2 26.37, A3 800.69) is better than the calculated MSE by K-means (A1 1632.88, A2 2516.83, A3 3052.36) and ORC algorithm (A1 1311.88, A2 2138.64, A3 2470.51).

**Table 2. Best MSEs obtained.**

| Algorithm | A1 | A2 | A3 |
|---|---|---|---|
| K-means | 1632.88 | 2516.83 | 3052.36 |
| ORC | 1311.88 | 2138.64 | 2470.51 |
| Proposed method | 14.77 | 26.37 | 800.69 |

In summary, the above experimental results on A1, A2, and A3 datasets show that the proposed algorithm can identify outliers more successfully than existing algorithms.

# 6. CONCLUSION

In this paper, we have proposed to integrate outlier removal into clustering for nonparametric model estimation. The proposed method was also compared with the standard K-means without outlier removal, and a simple approach in which outlier removal precedes the actual clustering. The proposed method employs both clustering and outlier discovery to improve estimation of the centroids of the generative distribution.

The experimental results show that the proposed method can identify outliers more successfully than existing algorithms and indicate that our method has a lower error on datasets with overlapping clusters than the competing methods.

# 7. ACKNOWLEDGMENTS

We want to give our special thanks to Prof. Adel Abo El-Magd for his encouragement and support with the evaluation.

# 8. REFERENCES

[1] Williams, G., Baxter, R., He, H., Hawkins, S., and Gu, L.2002. A Comparative Study for RNN for Outlier Detection in Data Mining. In Proceedings of the 2nd IEEE International Conference on Data Mining, Maebashi City, Japan, pp.709.

[2] He,Z., Xu, X., and Deng,S. 2003. Discovering Cluster-based Local Outliers. Pattern Recognition Letters, vol.24, pp.1641-1650.

[3] Aggarwal, C., and Yu,P.2001. Outlier Detection for High Dimensional Data. In Proceedings of the ACM SIGMOD International Conference on Management of Data, vol.30, pp.37-46.

[4] Jaing, M., Tseng, S., and Su, C.2001. Two-phase Clustering Process for Outlier Detection. Pattern Recognition Letters, vol.22, pp.691-700.

[5] Taloba, A. I. 2008. Data Clustering Using Evolutionary Algorithms. Master thesis, Assiut University, Assiut,Egypt.

[6] Zhang, T.,Ramakrishnan, R., and Livny, M.1997. BIRCH: A new data clustering algorithm and its applications. Data Mining and Knowledge Discovery, vol.1,pp.141-182.

[7] Ester, M.,Kriegel, H. P., Sander J., and Xu, X.1996. A density-based algorithm for discovering clusters in large spatial databases with noise. In:2nd International Conference on Knowledge Discovery and Data Mining, pp.226-231.

[8] Guha, S.,Rastogi, R., and Shim, K.1999. A robust clustering algorithm for categorical attributes. In 15th International Conference on Data Engineering, pp.512-521.

[9] Pamula, R., Deka, J.K., Nandi, S. 2011. An Outlier Detection Method Based on Clustering. Emerging Applications of Information Technology (EAIT), pp. 253 – 256.

[10] Al-Zoubi, M., Al-Dahoud, A. and Yahya, A.A. 2010. New Outlier Detection Method Based on Fuzzy Clustering, WSEAS Transactions on Information Science and Applications, pp.681-690.

[11] Murugavel, P., and Punithavalli, M. 2011. Improved Hybrid Clustering and Distance-based Technique for Outlier Removal, International Journal on Computer Science and Engineering (IJCSE).

[12] Karmaker, A. and Rahman, S. 2009 Outlier Detection in Spatial Databases Using Clustering Data Mining, Sixth International Conference on Information Technology: New Generations, pp.1657-1658.

[13] Loureiro,A., Torgo, L. and Soares, C. 2004. Outlier Detection using Clustering Methods: a Data Cleaning Application, in Proceedings of KDNet Symposium on Knowledge-based Systems for the Public Sector. Bonn, Germany.

[14] Niu, K., Huang, C., Zhang, S., and Chen, J. 2007. ODDC: Outlier Detection Using Distance Distribution Clustering, T. Washio et al. (Eds.): PAKDD 2007 Workshops, Lecture Notes in Artificial Intelligence (LNAI) 4819, pp. 332–343.

[15] Hautamaki, V., Karkkainen, I., and Franti, P.2004. Outlier detection using knearestneighbour graph. In 17th International Conference on Pattern Recognition (ICPR 2004), Cambridge, United Kingdom, pp.430-433.

[16] Hautamaki,V.Cherednichenko, S.,Karkkainen, I.,Kinnunen, T.,and Franti, P.2005. Improving K-Means by Outlier Removal. In: SCIA 2005, pp.978-987.

[17] Virmajoki, O. 2004. Pairwise Nearest Neighbor Method Revisited. PhD thesis, University of Joensuu, Joensuu, Finland.